# Predicting Progression of Age-related Macular Degeneration from Fundus Images using Deep Learning

Boris Babenko, PhD[1]*
Siva Balasubramanian, MD, PhD[1]*
Katy E. Blumer, BS[1]
Greg S. Corrado, PhD[1]
Lily Peng, MD, PhD[1]
Dale R. Webster, PhD[1]
Naama Hammel, MD[1]**
Avinash V. Varadarajan, MS[1]**

*Equal contribution
**Equal contribution

Affiliations:
[1]Google AI Healthcare, Google LLC, Mountain View, CA, USA

Corresponding author:
Naama Hammel, MD
Google AI Healthcare
1600 Amphitheatre Parkway
Mountain View, CA 94043
nhammel@google.com

## Abstract

**Background:** Patients with neovascular age-related macular degeneration (AMD) can avoid vision loss via certain therapy. However, methods to predict the progression to neovascular age-related macular degeneration (nvAMD) are lacking.

**Purpose:** To develop and validate a deep learning (DL) algorithm to predict 1-year progression of eyes with no, early, or intermediate AMD to nvAMD, using color fundus photographs (CFP).

**Design:** Development and validation of a DL algorithm.

**Methods:** We trained a DL algorithm to predict 1-year progression to nvAMD, and used 10-fold cross-validation to evaluate this approach on two groups of eyes in the Age-Related Eye Disease Study (AREDS): none/early/intermediate AMD, and intermediate AMD (iAMD) only. We compared the DL algorithm to the manually graded 4-category and 9-step scales in the AREDS dataset.

**Main outcome measures:** Performance of the DL algorithm was evaluated using the sensitivity at 80% specificity for progression to nvAMD.

**Results**: The DL algorithm's sensitivity for predicting progression to nvAMD from none/early/iAMD (78±6%) was higher than manual grades from the 9-step scale (67±8%) or the 4-category scale (48±3%). For predicting progression specifically from iAMD, the DL algorithm's sensitivity (57±6%) was also higher compared to the 9-step grades (36±8%) and the 4-category grades (20±0%).

**Conclusions**: Our DL algorithm performed better in predicting progression to nvAMD than manual grades. Future investigations are required to test the application of this DL algorithm in a real-world clinical setting.

# Introduction

Age-related macular degeneration (AMD) is the leading cause of irreversible vision loss among people aged 50 years or older in the industrial world.[1] Due to a rapid growth in the elderly population, the number of individuals visually impaired from AMD is expected to increase substantially. It is estimated that 187 million people worldwide currently suffer from AMD, a statistic projected to reach 196 million in 2020, and 288 million in 2040.[2]

The diagnosis of AMD is based on the clinical presentation, with drusen being the hallmark of early AMD.[3] Intermediate AMD (iAMD) is characterized by the presence of numerous intermediate drusen, at least one large druse, or non-central geographic atrophy (GA).[3] An individual with iAMD is at increased risk of progression to late-stage AMD.[3] Late (or advanced) AMD can be classified into dry or wet forms. Advanced dry AMD is characterized by central geographic atrophy (CGA) involving the center of the fovea, while wet (or neovascular) AMD (nvAMD), is characterized by the presence of choroidal neovascular membranes that lead to severe central vision loss.[4] Although nvAMD only accounts for approximately 10%–20% of the overall incidence of AMD, this subtype is responsible for 90% of cases of severe vision loss or legal blindness.[5,6] In the clinic, nvAMD is routinely managed using anti-vascular endothelial growth factor (anti-VEGF) agents.[3] Thus, identifying patients at greater risk of progressing to nvAMD can potentially help clinicians follow these patients closely and initiate treatment in a timely manner.

To study the clinical course, prognosis, and risk factors of AMD and age-related cataract, a long term, multicenter, prospective, randomized clinical trial, the Age Related Eye Disease Study (AREDS) was performed.[7] Over the course of the study, several grading scales were developed to determine the severity and predict the risk of progression of AMD, such as the 4-category classification scale (AREDS report number 6),[8] and the 9-step severity scale

(AREDS report number 17).[9] The 4-category scale is a descriptive scale designed to standardize grading of color fundus photographs (CFP) for AMD and is widely accepted in clinical practice for AMD classification.[3] The 9-step scale was developed to help determine risk of progression to advanced AMD within 5 years. Application of these classification systems in a clinic setting, in particular the detailed 9-step scale for disease progression, requires manual grading of fundoscopic changes of AMD by trained medical specialists or highly skilled graders, a task that is subjective, tedious and time consuming.

A potential solution to reduce workload and improve consistency of risk stratification is deep learning (DL).[10] DL has been applied to produce highly accurate algorithms that can detect eye conditions such as diabetic retinopathy (DR), AMD and glaucoma with accuracy comparable to human experts.[11–16] The use of deep learning based algorithms have been further shown to assist eye care providers with DR screening efforts.[17] However, most recent work on applying DL to AMD focuses on algorithms for image grading, such as assigning an AMD grade for a given CFP[18] or spectral domain optical coherence tomography (SD-OCT) images,[19] rather than prognosis: predicting progression to nvAMD.

In this study, we developed and validated a DL algorithm using CFP from the AREDS database to identify patients at high risk of progressing to nvAMD within 1-year. We focused on two groups: the first group consisted of eyes with either no, early, or iAMD (none/early/iAMD group), and the second group consisted only of eyes with iAMD (iAMD group).

## Methods

### *Datasets*

The dataset used for the analyses described in this manuscript was obtained from the AREDS database found at https://dbgap.ncbi.nlm.nih.gov/aa/wga.cgi?page=login through dbGaP accession number [phs000001.v3.p1.c1]. The study protocol was approved by each clinical center's institutional review board.[20]

### Grades

We used grades from two AMD grading scales: the 4-category classification scale and the 9-step severity scale. The 9-step scale represents changes seen in early to iAMD, and an additional three steps capture late-AMD: CGA (step 10), nvAMD (step 11) and CGA with nvAMD (step 12). Since the 4-category scale considers both CGA and nvAMD as advanced AMD, we split the 4th category into two subcategories: CGA only, and nvAMD (with or without CGA). Distribution of grades across all stereo-pairs are shown in **Table 1**.

### Grades reversals

Since CFP from different visits for the same patient were graded independently in the AREDS protocol, there were some cases where an eye was graded as having advanced AMD in one visit and less than advanced AMD in subsequent visits. However, since such reversals are merely an artifact of the graders not having access to prior diagnoses, we excluded the CFP where these reversals were observed, as well as subsequent CFP, when developing and evaluating our algorithms.

### Data pre-processing

We used an automated text detection system to discard images that contain text, to ensure that the images were completely de-identified. We also discarded images for which we were unable to automatically detect the circular boundary (approximately 2% of the images). The remaining images were resized to be 587 pixels in each dimension. **Table 1** contains the statistics of the final dataset used in all experiments.

*Algorithm development*

We developed a DL algorithm that takes stereo pairs of CFP of a single eye as input. We trained models in a supervised fashion, feeding a binary label corresponding to whether the eye progressed to nvAMD within 1-year after the CFP was captured. The architecture of the model was based on Inception-v3.[21] The original Inception-v3 architecture[21] takes a single red, green, blue (RGB) image as input. To adapt this architecture to take a stereo pair as input, we used a "late fusion" approach: the left and right stereo CFP were each fed into separate Inception-v3 networks (with shared weights), and the outputs (post-softmax) were averaged. Since our network had a large number of parameters (22 million), early stopping[22] was used to terminate training before convergence. To speed up training, the network was initialized using parameters from a network pre-trained to classify objects in the ImageNet dataset.[23] The model was trained using TensorFlow,[24] following a procedure similar to that previously described by Krause et al.[12]

*Algorithm evaluation*

We used 10-fold cross validation to evaluate our DL algorithm and compare it with the 4-category and 9-step manual grades' performance in predicting progression to nvAMD within 1-year. Specifically, we split the entire AREDS dataset randomly into 10 approximately equal-sized partitions that were mutually exclusive with respect to patients. We then used a standard 10-fold cross-validation approach: developing a model on 9 of the folds, and evaluating the models on the single held-out fold, repeating 10 times such that each of the folds was used for evaluation once. In each iteration, within the 9 folds used for development, we used one fold for tuning the model (e.g. determining early stopping criteria), and the other 8 for training.

We computed the following metrics: the sensitivity at 80% specificity, receiver operating characteristic (ROC) curves, and area under the ROC curve (AUC). The 80% specificity operating point was selected using the respective tuning fold, and the specificity and sensitivity on the held out "testing" folds was computed.

Both eyes of each patient were included in the dataset and treated as independent data points. However, we observed that eyes that did not progress had, on average, more visits in the dataset than eyes that did progress. Hence, treating the *visits* as independent data points may introduce a bias. To avoid such bias we employed the following strategy: we sampled one visit per eye 100 times, computed metrics for each sample and then computed an average to produce metrics for each testing fold. Finally, we computed the averages and standard deviations of metrics across the 10 folds.

We computed the metrics on eyes with none/early/iAMD in the test set and also just on eyes with iAMD, and evaluated 4-category and 9-step grades on the same 10 folds for purposes of comparison, computing all the same metrics as described above (**Figure 1**).

## Results

Among the none/early/iAMD group, the DL algorithm had a sensitivity of 78±6%, and the manual grades from the 4-category scale and 9-step scale had sensitivities of 48±3% and 67±8%, respectively. For the iAMD only group, the DL algorithm had a sensitivity of 57±6%, and the 4-category grades and 9-step grades had sensitivities of 20±0% and 36±8%, respectively (**Table 3 and Figure 2**).

To further illustrate how the DL model stratifies eyes by their risk of progression, we split the data points of each test fold into quartiles by their corresponding DL algorithm prediction and computed the actual progression rate in each quartile, and finally averaged these rates across the 10 folds. This visualization is shown in **Figure 3**. The eyes predicted by the algorithm to be in the highest quartile of progression risk had a 12.4% actual progression rate among none/early/iAMD eyes (compared to a baseline rate of 3.7%), and 17.2% progression rate among iAMD eyes (baseline of 7.2%). By contrast, the eyes in the lowest quartile had actual progression rates below 1%.

## Discussion

In this study we compared the performance of a DL algorithm with manual grades in predicting 1-year progression to nvAMD among two groups: eyes with none, early, or iAMD and eyes with iAMD. Our DL algorithm performed better than manual grades in predicting 1-year progression.

### *Performance of the algorithm*

On the task of predicting progression of None/Early/iAMD eyes to nvAMD within 1-year, we observe that our DL algorithm was more sensitive than manual grades from both grading scales, and that the 9-step grades outperform the 4-category grades. The latter observation was not surprising because the 9-step scale was explicitly developed to predict disease progression based on outcomes data, albeit focusing on the 5-year time point.[9] The superiority of the DL algorithm could be due to two factors: the potential inconsistency in manual grading, and the ability of the DL algorithm to detect novel features that are not captured in existing grading scales. We conducted additional experiments to help dissect these two factors, finding that the superior performance is due to both factors, but the former is likely more significant (more details in the Appendix).

On the iAMD group we observe that the DL algorithm was again more sensitive. Note that in this group, all the 4-category grades are identical, so the ROC curve is a diagonal line corresponding to random performance, and the sensitivity at 80% specificity is 20%. Although the DL algorithm outperforms the 9-step grades, we observe that the absolute performance numbers of both approaches are lower on the iAMD group than the none/early/iAMD group. This is likely because the none/early/iAMD group is a heterogeneous cohort including a number of low risk eyes (e.g. no AMD), and the ability to correctly identify these easier cases improves the algorithm's performance. By contrast, the iAMD group is a comparatively more homogenous cohort that is more difficult to risk stratify.

### *Related work*

Several studies have reported the use of DL algorithms to detect AMD using CFP. Burlina et al.[25] compared the performance of humans and DL algorithm in grading CFP to detect AMD. Peng et al.[26] and Grassmann et al.[18] developed DL algorithms for automated classification of AMD. Recently, Burlina et al.[27] trained a DL algorithm to predict the 9-step grades and also the associated known 5-year progression rates to advanced AMD from CFP. By contrast, our study aims to directly predict whether an eye will progress to nvAMD within 1-year via binary classification. Our approach has the advantage of allowing the DL algorithm to learn any features in the CFP that are predictive of progression, instead of restricting it to recapitulate the features associated with the 9-step grade. A recent study by Banerjee et al.[28] used a binary classification approach similar to our work, reporting an AUC of 0.68 in predicting fellow-eye progression to nvAMD using SD-OCT scans from two visits using a recurrent neural network. Of interest, SD-OCT has been reported to be highly sensitive in detecting nvAMD, though it may not fully replace CFP or fluorescein angiography (FA) in detecting various lesions in AMD.[29] Future investigations are required to build a DL algorithm using data captured from multimodal imaging wherein CFP, SD-OCT, or FA may complement each other in identifying different lesions or types of AMD missed by one or the other imaging technique.

*Limitations*

Our study is not without limitations. AREDS grades were based on assessment of CFP alone without the use of FA to validate the ground truth, in particular the nvAMD grades. The current model was developed using stereoscopic CFP and may not be applicable to two-dimensional images captured in screening programs. In addition, the AREDS cohort lacks ethnic diversity and consists mostly of Caucasian participants. Nevertheless, our DL algorithm could have potential implications in clinical practice for an automated diagnosis of AMD and tracking of progression to nvAMD.

*Conclusion*

In conclusion, our study predicted the 1-year progression to nvAMD for two groups of eyes in the AREDS dataset, those with either no, early, or iAMD, and those with iAMD. Predicting short term progression could help eye care providers identify patients who require a closer follow-up in order to detect nvAMD earlier, initiate treatment earlier, and potentially achieve better visual acuity outcomes. DL algorithms could also help enrich clinical trials or study cohorts with patients who eventually progress, thus aiding future investigations aimed at slowing or treating early or iAMD.

# Tables and Figures

| Table 1. Baseline characteristics of dataset | |
|---|---|
| **Patient demographics** | |
| No. of patients | 4,628 |
| No. of eyes | 9,221 |
| No. of stereo pairs | 71,772 |
| Median age at baseline in years (IQR) | 69 (65-73) |
| Females (%) | 40,329 (56%) |
| Median No. of visits (IQR) | 9 (6-11) |
| **Distribution of 4-category manual grades across all stereo pairs** | |
| None | 22,897 (32%) |
| Early | 16,547 (23%) |
| iAMD | 22,297 (31%) |
| CGA | 1,537 (2%) |
| nvAMD | 8,494 (12%) |
| **Distribution of 9-step severity scale manual grades across all stereo pairs** | |
| Step 1 | 26,471 (37%) |
| Step 2 | 7,744 (11%) |
| Step 3 | 3,326 (5%) |
| Step 4 | 6,023 (8%) |
| Step 5 | 3,905 (5%) |
| Step 6 | 4,971 (7%) |
| Step 7 | 4,417 (6%) |
| Step 8 | 3,518 (5%) |
| Step 9 | 1,355 (2%) |
| CGA (Step 10) | 1,537 (2%) |
| nvAMD (Step 11) | 8,029 (11%) |
| CGA+nvAMD (Step 12) | 465 (1%) |

Abbreviations: IQR, Interquartile range; iAMD, intermediate age-related macular degeneration; CGA, Central geographic atrophy; nvAMD, neovascular age-related macular degeneration;

| Table 2. Statistics for 1-year progression to nvAMD in the study dataset | | |
|---|---|---|
| | none/early/iAMD | iAMD |
| **No. of eyes** | 8,451 | 4,468 |
| **No. of visits** | 61,741 | 22,297 |
| **No. of visits with an available AMD grade at 1-year followup** | 41,569 | 14,731 |
| **No. of visits graded as nvAMD at 1-year followup** | 578 | 532 |
| **Adjusted* 1-year rate of progression to nvAMD per eye** | 3.7% | 7.2% |

Abbreviations: nvAMD, neovascular age-related macular degeneration; iAMD, intermediate age-related macular degeneration.

This table shows the number of eyes that fell into the none/early/iAMD and iAMD groups at any point in the trial, the total number of visits in each group, the subset of those visits that had a known outcome (i.e. whether the eye progresses to nvAMD within 1 year), the subset of those visits that progress, and a progression rate. *The progression rate is adjusted to avoid biases towards eyes with more visits (using the sampling strategy described in the *Algorithm evaluation* section in the Methods).

| | None/Early/iAMD | | | iAMD | | |
|---|---|---|---|---|---|---|
| | AUC* | Sensitivity (%) | Specificity (%) | AUC* | Sensitivity (%) | Specificity (%) |
| **4-category scale** | 0.78±0.2 | 48±3 | 80±1 | 0.50±0.00 | 20±0 | 80±0 |
| **9-step scale** | 0.83±0.03 | 67±8 | 80±2 | 0.67±0.05 | 36±8 | 80±2 |
| **Deep Learning** | 0.88±0.02 | 78±6 | 80±2 | 0.77±0.04 | 57±6 | 81±4 |

Table 3. Comparing performance of the deep-learning algorithm with AMD grades for predicting progression to nvAMD at 1-year

Abbreviations: AMD, age-related macular degeneration; nvAMD, neovascular age-related macular degeneration; iAMD, intermediate age-related macular degeneration;
*Results are presented as mean±std across the 10 cross validation folds.

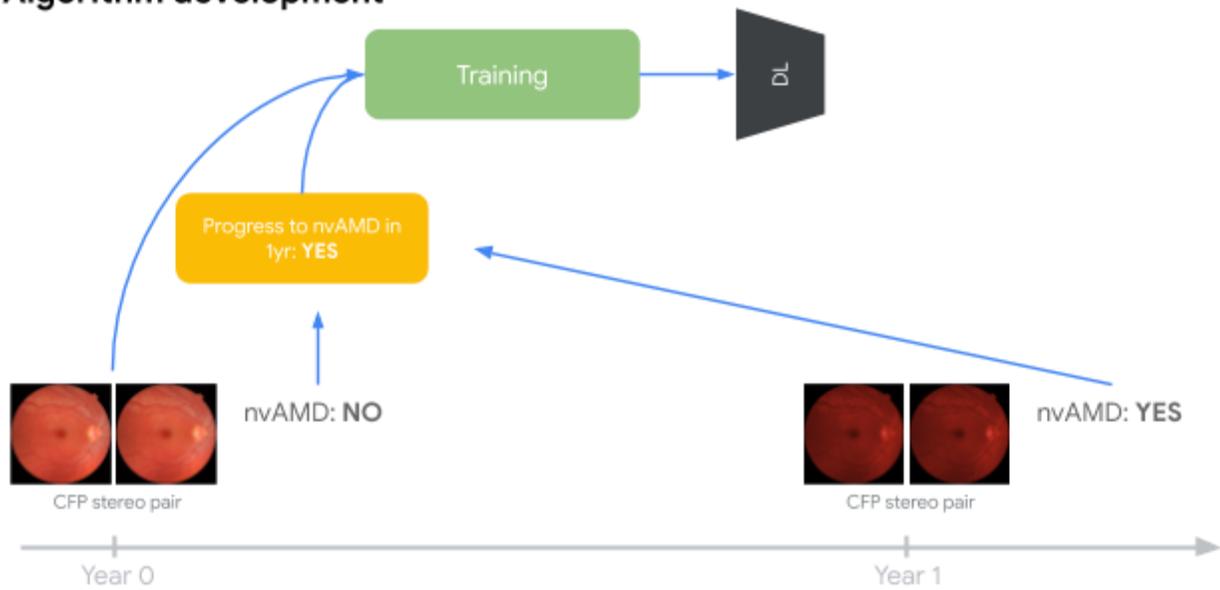

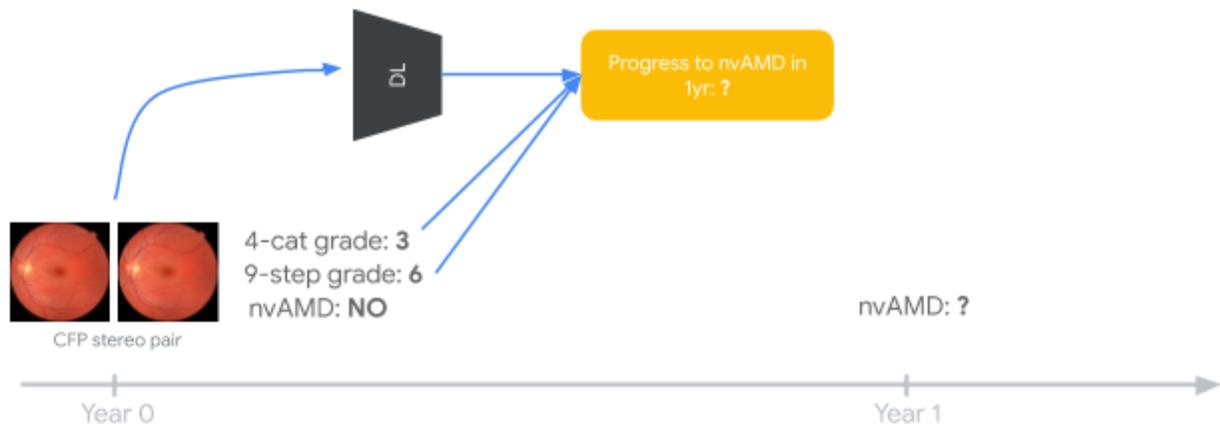

**Figure 1.** (Top) During training, we fed pairs of images and labels into the training algorithm, where labels correspond to whether the eye progresses to nvAMD within 1 year after the CFP is captured. (Bottom) During evaluation, we either feed the CFP into a trained DL model, or use the 4-category or 9-step grade to predict whether the eye will progress to nvAMD within 1 year. (fundus photos source: Häggström, Mikael (2014). "Medical gallery of Mikael Häggström 2014". *WikiJournal of Medicine* **1** (2). DOI:10.15347/wjm/2014.008. ISSN 2002-4436. Public Domain)

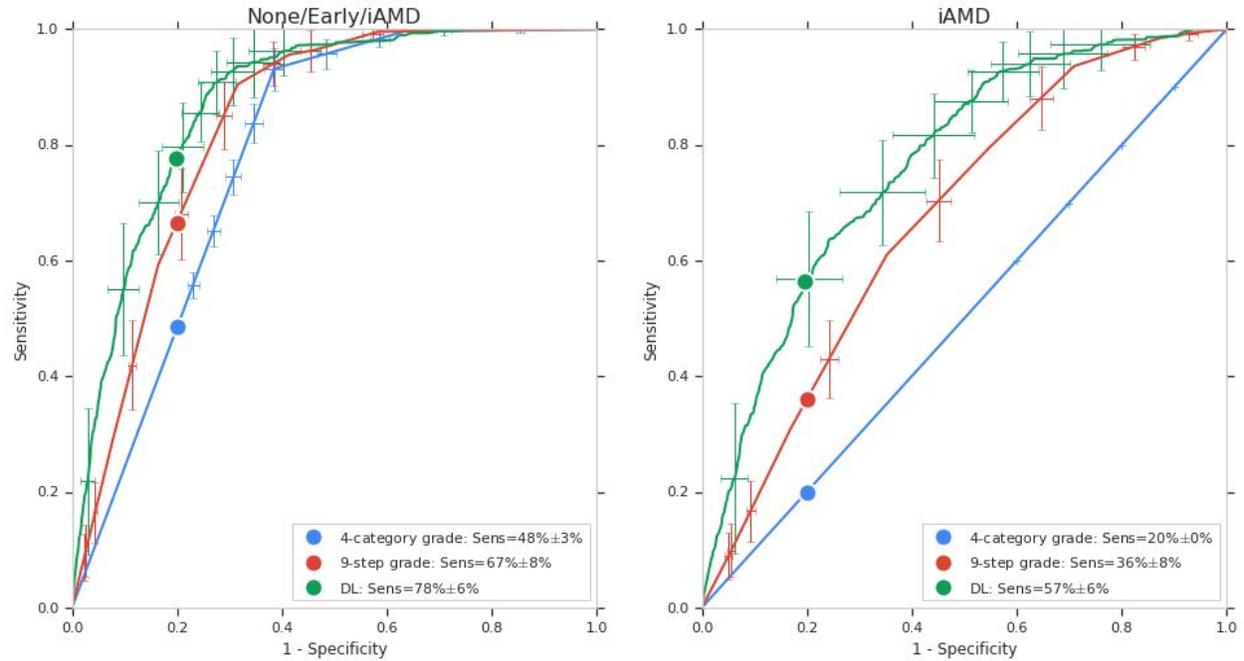

**Figure 2.** Receiver operating characteristic curve (ROC) of manual grades and deep learning (DL) algorithm for predicting progression to neovascular AMD from (left) none/early/intermediate AMD and (right) intermediate AMD (iAMD). Error bars indicate standard deviation across the 10 cross validation folds, and are randomly sampled for visualization purposes. ROC plots for all folds are available in the Appendix.

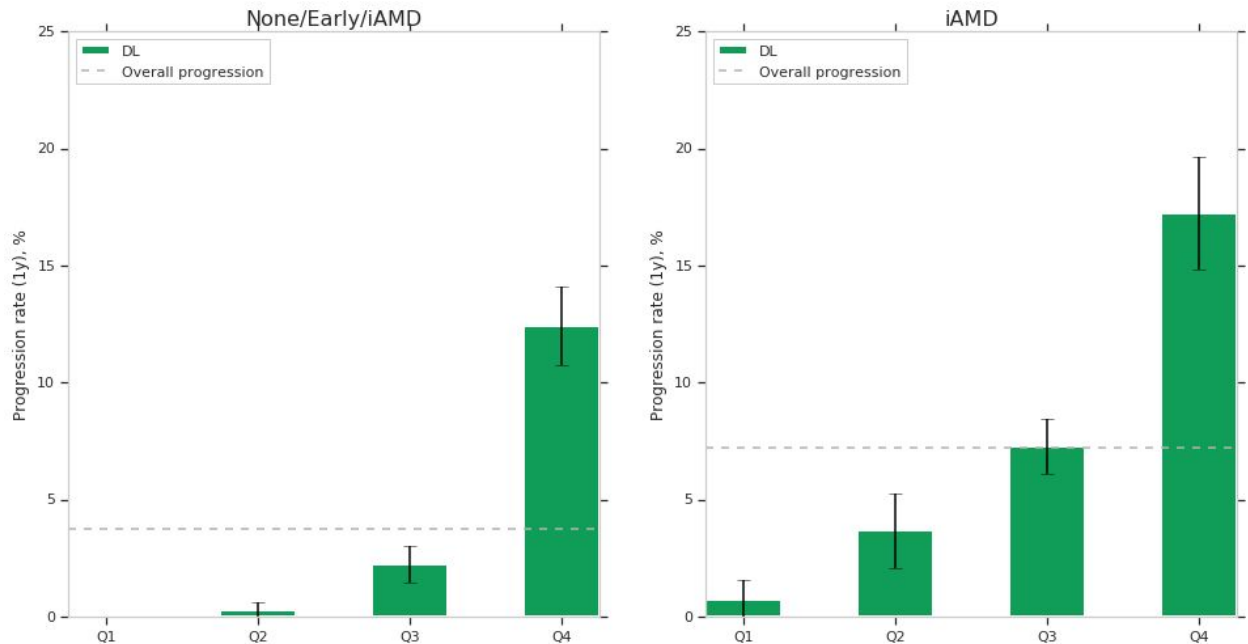

**Figure 3.** 1-year rate of progression to nvAMD from (left) none/early/intermediate AMD and (right) intermediate AMD (iAMD) for the 4 quartiles using our proposed DL algorithm. Error bars indicate standard deviation across the 10 cross-validation folds.

**Acknowledgements**

The dataset used for the analyses described in this manuscript were obtained from the NEI Study of Age-Related Macular Degeneration (NEI-AMD) Database found at [https://www.ncbi.nlm.nih.gov/projects/gap/cgi-bin/study.cgi?study_id=phs000001.v3.p1] through dbGaP accession number [phs000001.v3.p1.c1]. Funding support for NEI-AMD was provided by the National Eye Institute (N01-EY-0-2127). We would like to thank NEI-AMD participants and the NEI-AMD Research Group for their valuable contribution to this research.

## Appendix

In this appendix, we explore several hypotheses for why the DL algorithm presented in the main text outperforms manual grading along the two grading scales: Is the algorithm picking up novel features that the grading scales do not take into account? Is the algorithm using the same features as the grading scales, but doing so more consistently than manual grading and thus more accurately predicting progression? Or is it simply more accurate because it is not constrained to one of a small number of discrete grades?

To help answer these questions, we explored "two-phase" DL algorithms (**Figure A1**). We shall refer to the DL algorithm presented in the main text as "end-to-end" DL algorithm. In the two-phase setup, a DL model predicts the AMD grade of the original CFP stereo pair taken in the current visit (i.e. the "current" AMD grade), and then uses the *predicted* grade to predict whether the eye will progress to nvAMD in 1-year. We explored this type of approach for both 4-category and 9-step grading scales. Furthermore, because a DL algorithm trained to predict the current AMD grade outputs a "likelihood" *distribution* over the steps in the respective grading scales, we also explored a two-phase approach where the predicted distribution over the respective grading scale are fed into a logistic regression (LR) model that is trained to predict whether the eye will progress to nvAMD within 1-year. We will refer to these two-phase algorithms as "two-phase (mode)" that predicts a single grade (i.e., the most likely grade, or "mode"), and "two-phase (LR)" that predicts both a distribution over the grading scale as well the final progression prediction.

The performances of two-phase algorithms, end-to-end DL algorithm and manual grades from the two grading scales among both none/early/iAMD and iAMD groups are shown in **Figure A2** and **Table A1**, and progression rates are shown for the algorithms that output continuous scores in **Figure A3.** For the 4-category two-phase models we observed that on both none/early/iAMD and iAMD groups, the LR version achieved higher sensitivities than the mode version (74±6% and 50±6% vs 54±5% and 31±5%, respectively). Similarly, for the 9-step two-phase models, we observed that the LR version achieved higher sensitivities as well (76±5% and 46±8% vs 76±6% and 40±7%), but the difference is much smaller compared to what we see for the 4-category two-phase models. This suggests that the coarseness of the grading scale affects its utility to accurately predict progression; the 9-step scale is fine-grained, such that even a single, discrete grade carries enough information to predict progression reasonably well. Furthermore, we observe that the two-phase LR algorithms for both the 4-category and 9-step grading scales achieved similar performance. That is, despite having a different granularity, the algorithm-captured likelihood distribution of the two grading scales seem to capture similar information regarding progression.

We also observed that the two-phase (mode) models for both grading scales perform better than their corresponding manual grades. For example, the 9-step two-phase (mode) model achieved a sensitivity of 40±7% for the iAMD group, while the 9-step manual grades achieved a sensitivity of 36±8%. This may be due to the model being more consistent than human graders, which then enables better progression prediction.

Finally, we observed that the LR versions of both two-phase models produce results that are slightly below that of the end-to-end DL algorithm on both groups, suggesting that an end-to-end DL algorithm may be picking up on some features or signals that neither grading scale captures.

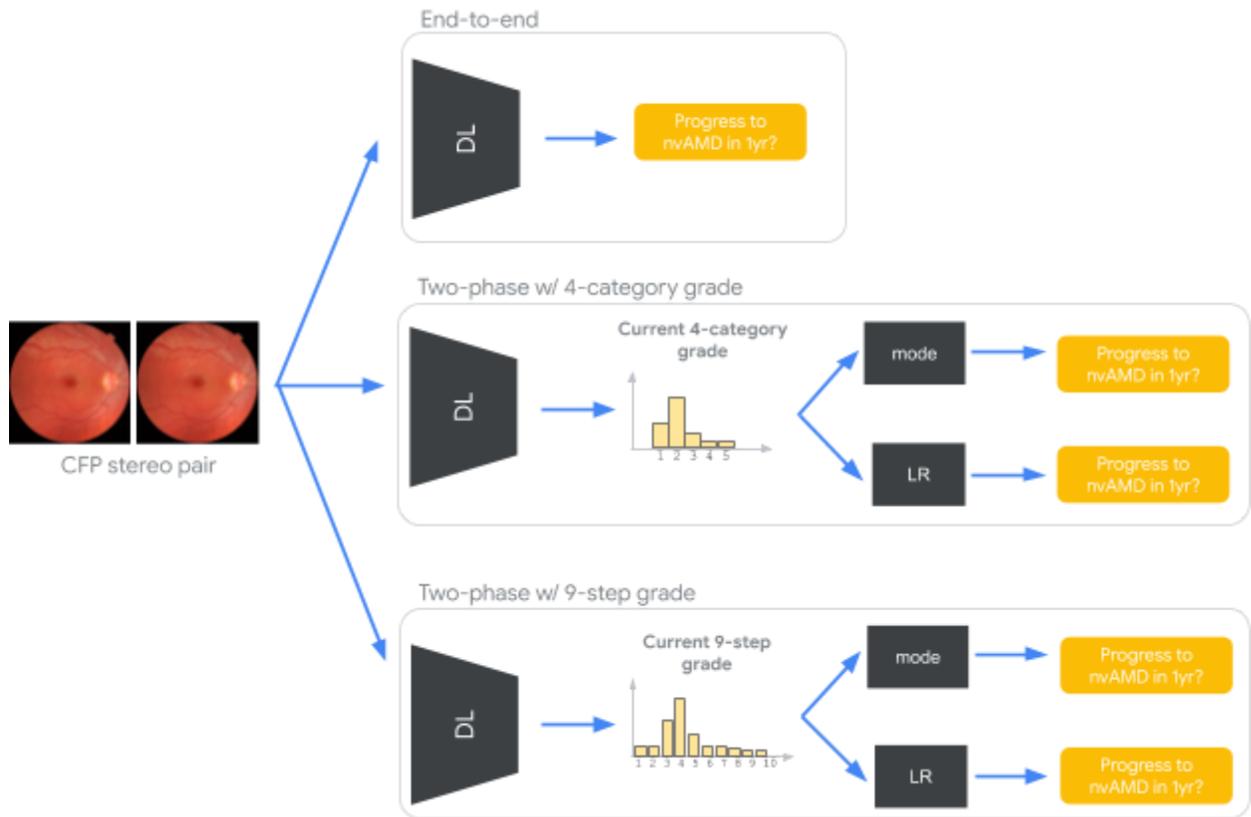

**Figure A1**. Schematic diagram showing the DL algorithms developed to analyze fundus stereo pairs.

| | | None/Early/iAMD | | | iAMD | | |
|---|---|---|---|---|---|---|---|
| | | AUC* | Sensitivity (%) | Specificity (%) | AUC* | Sensitivity (%) | Specificity (%) |
| **4 category** | Manual | 0.78±0.2 | 48±3 | 80±1 | 0.50±0.00 | 20±0 | 80±0 |
| | Two-phase (mode) | 0.80±0.2 | 54±5 | 80±1 | 0.61±0.03 | 31±5 | 80±2 |
| | Two-phase (LR) | 0.86±0.2 | 74±6 | 80±2 | 0.74±0.04 | 50±6 | 80±3 |
| **9 step** | Manual | 0.83±0.03 | 67±8 | 80±2 | 0.67±0.05 | 36±8 | 80±2 |
| | Two-phase (mode) | 0.85±0.2 | 76±6 | 80±2 | 0.71±0.02 | 40±7 | 80±2 |
| | Two-phase (LR) | 0.86±0.2 | 76±5 | 80±2 | 0.74±0.04 | 46±8 | 80±3 |
| **DL (end-to-end)** | | 0.88±0.02 | 78±6 | 80±2 | 0.77±0.04 | 57±6 | 81±4 |

Table A1. Comparing performance of two-phase models for predicting progression to nvAMD at 1-year

Abbreviations: nvAMD, Neovascular Age-related Macular Degeneration; iAMD, Intermediate Age-related Macular Degeneration

*Results are presented as mean±std across the 10 cross-validation folds

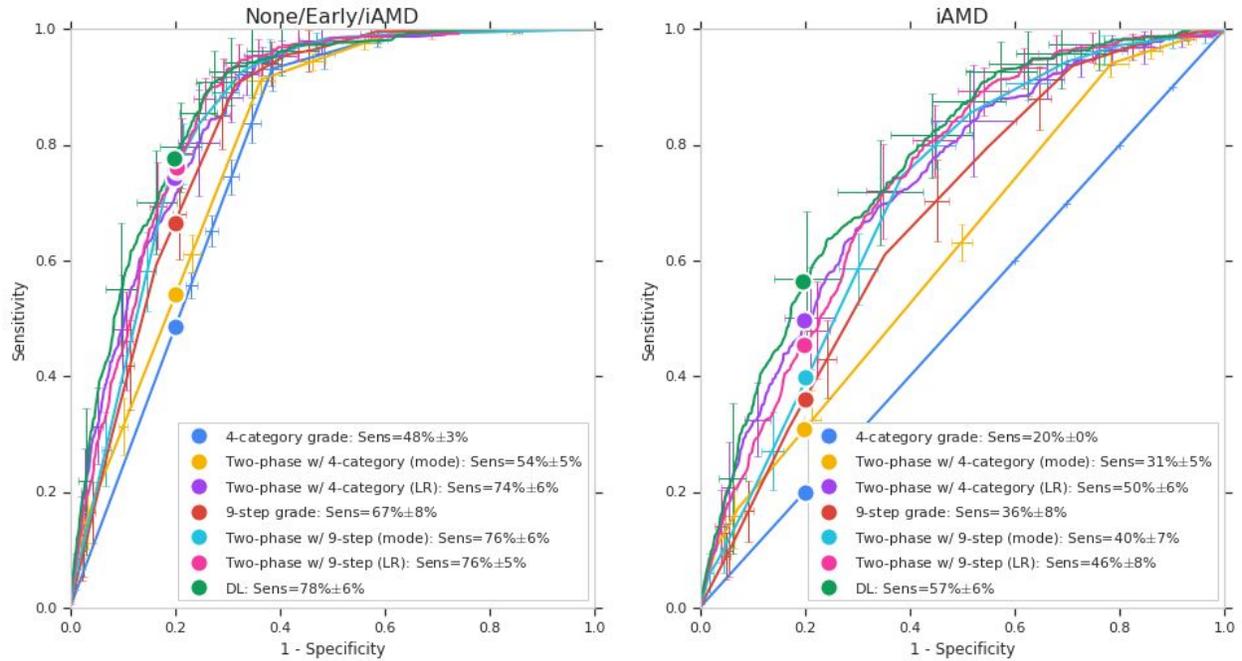

**Figure A2.** Receiver operating characteristic curve (ROC) analysis comparing two-phase algorithms with the end-to-end DL algorithm and manual grades from the two grading scales on (left) none/early/intermediate AMD and (right) intermediate AMD. Error bars indicate standard deviation across the 10 cross validation folds, and are randomly sampled for visualization purposes.

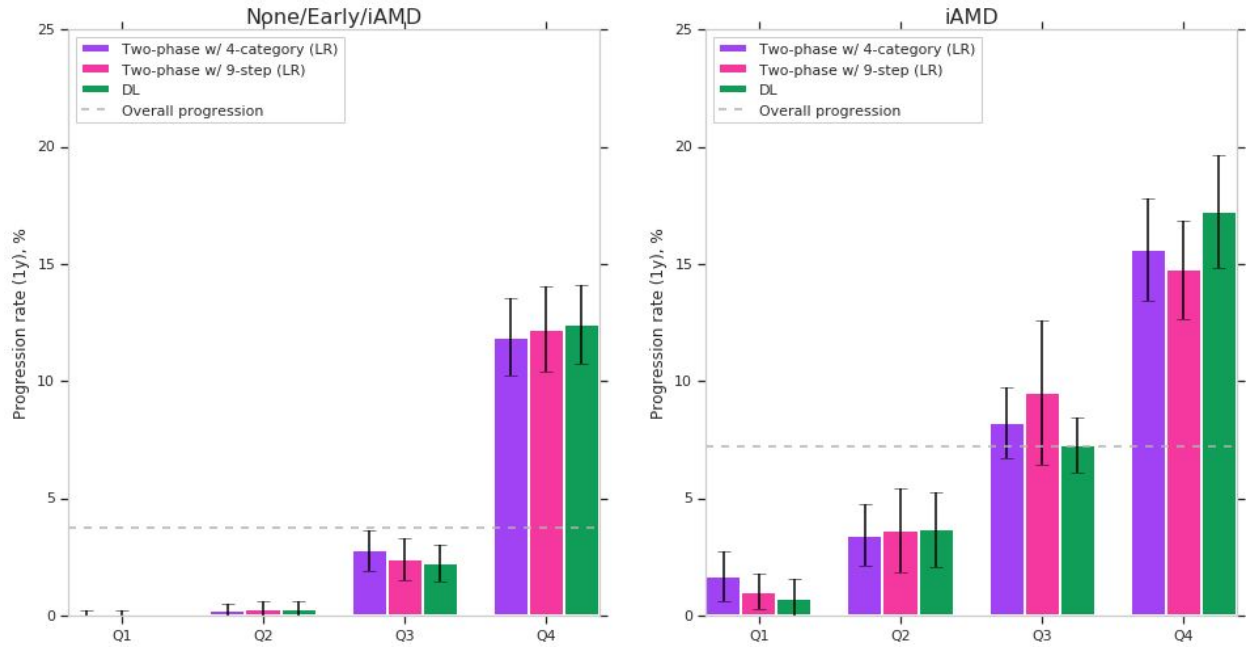

**Figure A3.** Progression to nvAMD from (left) none/early/iAMD and (right) iAMD for quartiles using DL scores, and scores from the (LR) versions of the two-phase algorithms for both grading scales.

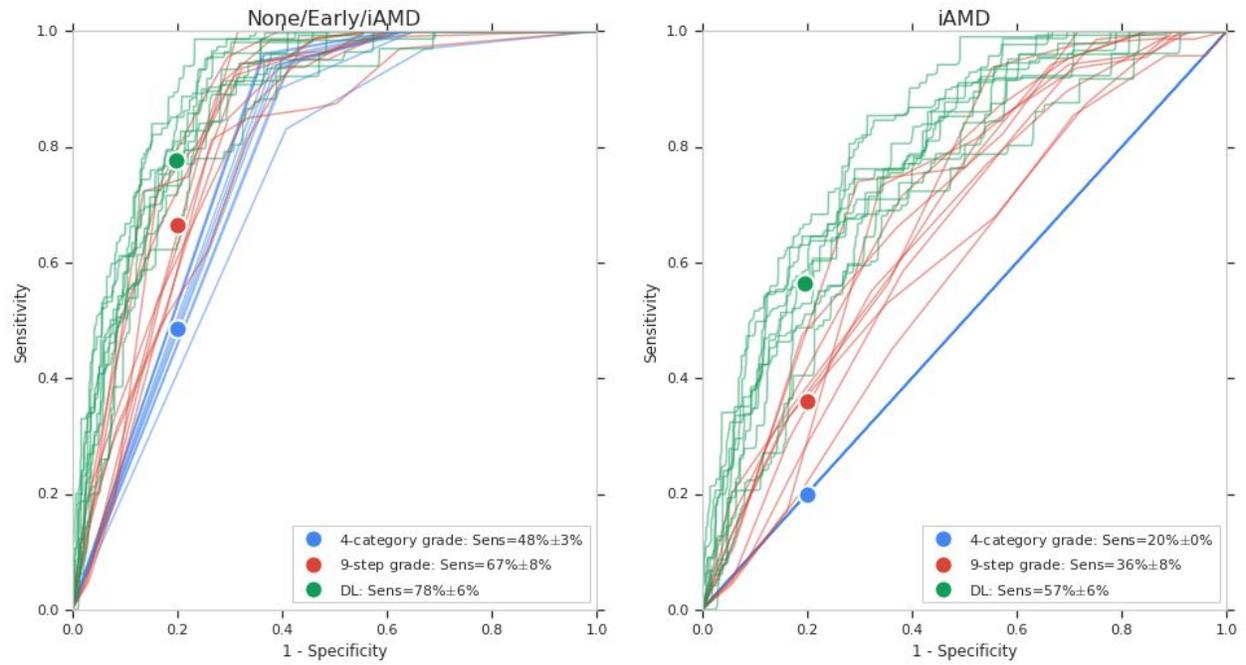

**Figure A4.** Alternate version of **Figure 2** in the main text. ROC plots showing curves for each cross fold rather than standard deviation error bars.